\documentclass[preprint,turnofflinenumbers,]{JASA-EL}
\usepackage{amsmath} 
\usepackage{gensymb} 
\usepackage{tipa}
\usepackage{booktabs}
\usepackage{amsmath}
\newcommand{\mbeq}{\overset{!}{=}}  

\begin{document}

\title{Dynamical model parameters from ultrasound tongue kinematics}

\author{Sam Kirkham}
\email{s.kirkham@lancaster.ac.uk}
\affiliation{Phonetics Laboratory, Lancaster University, United Kingdom}

\author{Patrycja Strycharczuk}
\email{patrycja.strycharczuk@manchester.ac.uk}
\affiliation{Linguistics and English Language, University of Manchester, United Kingdom}

\date{\today}

\begin{abstract}
The control of speech can be modelled as a dynamical system in which articulators are driven toward target positions. These models are typically evaluated using fleshpoint data, such as electromagnetic articulography (EMA), but recent methodological advances make ultrasound imaging a promising alternative. We evaluate whether the parameters of a linear harmonic oscillator can be reliably estimated from ultrasound tongue kinematics and compare these with parameters estimated from simultaneously-recorded EMA data. We find that ultrasound and EMA yield comparable dynamical parameters, while mandibular short tendon tracking also adequately captures jaw motion. This supports using ultrasound kinematics to evaluate dynamical articulatory models.
\end{abstract}

\maketitle

\section{Introduction}
\label{sec:introduction}

A major goal in the study of speech communication is understanding the nature of articulatory control. A common approach is to cast this problem in terms of a dynamical system with point attractor dynamics, where a small number of parameters drive the vocal tract to a stable equilibrium position \cite{fowler1980, saltzman-munhall1989, browman-goldstein1986, gafos2006, tilsen2016}. A standard model in this framework is the linear harmonic oscillator,

\begin{equation}
m\ddot{x} + b\dot{x} + kx = 0
\label{eq:sm89_act}
\end{equation}

where $m$ is mass (typically $m = 1$), $k$ is a stiffness coefficient, and $b$ is a damping coefficient, usually set to critically damped $b = 2\sqrt{mk}$. Gestural activation can be governed by step activation, with gestural parameters changing instantaneously at the point of activation and remaining constant over the activation interval.

In this study we focus on whether the parameters of a linear harmonic oscillator can be estimated from ultrasound tongue imaging data, which we compare with the more common method of fitting to electromagnetic articulography (EMA) data. A major barrier to this goal is that the linear harmonic oscillator is known to be a poor fit to empirical articulatory trajectories, as it predicts overly short time-to-peak velocity, meaning that it is inappropriate for evaluating how the model can fit different data modalities. There are three common solutions to this issue. The first allows gestural activation to vary over time \cite{byrd-saltzman1998}, which adds extrinsic complexity to the model. The second is a nonlinear model, such as adding a cubic term to the linear model \cite{sorensen-gafos2016, kirkham2025}, or novel nonlinear models \citep{stern-shaw2025}. The third is to abandon oscillatory models and develop new time-dependent (i.e. non-autonomous) models \citep{elie-etal2023}. All three approaches add significant complexity, but we take an alternative route, which is to retain the simple linear oscillator with step activation, but simply relax the critical damping constraint. This allows for a simple autonomous model that is generally more accurate than critically damped models \cite{kirkham2024, kirkham2025b}. We note that all of the above models focus only on piecewise dynamics, such as the movement between articulatory targets, so our decision to relax critical damping only adds a small level of complexity compared with nonlinear or non-autonomous models.

An important aspect of adjudicating between different models is evaluating their fit to empirical data, which allows us to establish prospective parameters for articulatory control. In other words: given an empirical articulatory trajectory, which model parameter values would be required to reproduce its dynamics? To date, the majority of dynamical articulatory model development has focused on fleshpoint tracking data, such as X-ray microbeam and EMA \cite{iskarous2016, elie-etal2023, kirkham2025b, stern-shaw2025}, with some applications to MRI data \cite{lammert-etal2013b}. Such data has a number of shortcomings, including limited information on the tongue posterior (EMA), invasive data collection (EMA, MRI), and limited portability (EMA, MRI). Ultrasound imaging largely overcomes these issues and provides good imaging of the tongue, as well as hyoid and mandibular short tendon \cite{wrench-balch-tomes2022}, but suffers from lower frame rates and noisy images. Despite this, recent advances suggest it is possible to derive kinematics from ultrasound, either via tracking manually-identified fanlines \cite{strycharczuk-scobbie2015} or anatomically-defined landmarks using deep learning \cite{wrench-balch-tomes2022}. For example, \citet{wrench-balch-tomes2022} trained a deep learning model on human-labelled data, where anatomically-defined landmarks were placed along the tongue. The accuracy of landmarking was comparable to between-human differences, allowing for automated frame-to-frame tracking of fleshpoint-like trajectory data, which also showed reasonable agreement with EMA. 

The above suggests that ultrasound is a candidate for estimating dynamical model parameters from data. This would be a substantial step forwards for evaluating dynamical models, as ultrasound is cheaper, less invasive, and provides richer information about lingual motion. It stands to reason that being able to accurately estimate dynamical parameters from ultrasound would open up a new range of applications for fieldwork and clinical data, which would facilitate model evaluation across more diverse samples and languages. In this study, we compare task dynamic parameters derived from simultaneous EMA and ultrasound data during vowel production. We focus on estimating the parameters of an undamped linear harmonic oscillator (i.e. Equation \ref{eq:sm89_act} but without the critical damping constraint). We use this model because it is simple and has known characteristics, which makes it an attractive case study for comparing model parameters estimated from ultrasound and EMA data. We expect that the same principles should apply to more complex models, but we use the simple model to establish a straightforward comparison without too many degrees-of-freedom.

\section{Methods}

\subsection{Speakers and stimuli}

The data set comprises simultaneous electromagnetic articulography and ultrasound tongue imaging data, which was recorded concurrently from six female speakers of Northern Anglo British English. The materials comprised the full set of British English vowels in /bV/ and /bVd/ contexts in two carrier phrases: \textit{She said X} and \textit{She said X eagerly}. Each speaker produced four repetitions of 29 words in two carrier phrases, except for one speaker who produced five repetitions. We excluded some blocks from two speakers due to excessive ultrasound probe movement. In total, we analyse 1095 tokens.

\subsection{Instrumentation}

EMA data were recorded using a Carstens AG501, with sensors placed on the tongue tip, tongue mid, tongue dorsum, upper/lower lip and lower teeth, with reference sensors located on the maxilla, nasion, and mastoids. The EMA data were recorded at 1250 Hz, filtered using a 50 Hz low-pass Kaiser-windowed filter (5 Hz for reference sensors), head corrected, and rotated to the occlusal plane. Ultrasound data were recorded in Articulate Assistant Advanced \cite{aaa2022} at $\sim$81 Hz using a Telemed MicrUS scanner with a 20-mm radius, 64-element, 2-MHz probe. The ultrasound probe was stabilised using a headset \cite{spreafico-eta2018}. Audio was recorded at 48 kHz using a Beyerdynamic Opus 55 microphone and pre-amplified using a Grace Designs m101 preamplifier. Audio, EMA and ultrasound data were time synchronized by aligning a TTL pulse that was triggered at the time of each prompt presentation and recorded onto each system. For further details of temporal synchronization and analysis of probe motion see \citet{kirkham-etal2023}.

\subsection{Data processing}

\begin{figure}[t]
  \centering
  \includegraphics[scale=0.72]{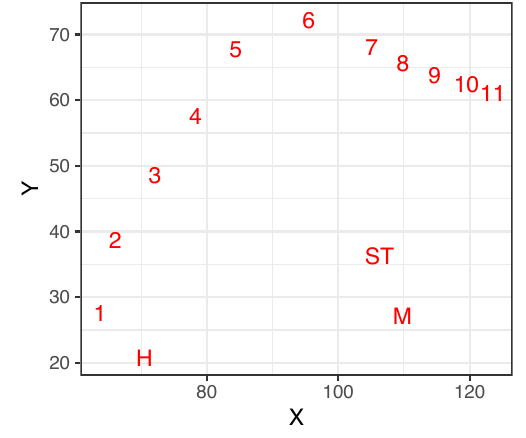}
  \includegraphics[scale=0.58]{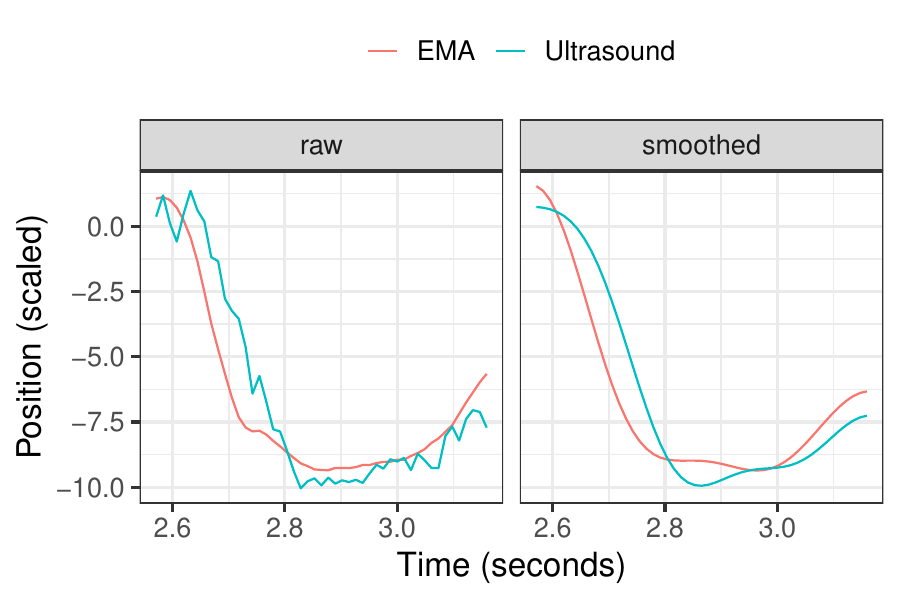}
  \caption{Left: Location of DLC knots estimated for each ultrasound frame (knot 1 is tongue root, knot 11 is tongue tip, H is hyoid, M is mandible, ST is short tendon). Right: Raw and smoothed data for TD horizontal position from EMA and ultrasound in the word \textit{bar}.}
  \label{fig:dlc_smooth}
\end{figure}

Acoustic data were forced-aligned using Montreal Forced Aligner \cite{mcauliffe-etal2017} and subsequently hand-corrected, which was used to segment the articulatory data based on the labelled CV interval. Anatomical landmarks in the ultrasound images were tracked in each frame using DeepLabCub \cite{mathis-etal2018}, which is a deep learning algorithm for markerless pose-estimation. We specifically used a pre-trained tongue model with 11 points at anatomical landmarks along the tongue, as well as points corresponding to the short tendon and hyoid bone \cite{wrench-balch-tomes2022}. Figure \ref{fig:dlc_smooth} (left) shows that knot 1 is located at the tongue root and knot 11 is located at the tongue tip, with all knots specified for x/y dimensions \cite{wrench-balch-tomes2022, strycharczuk-etal2025}. Knots were exported as Cartesian coordinates (in mm) and rotated parallel to the occlusal plane (estimated using a bite plate recording for each speaker). The EMA data were downsampled to the ultrasound frame rate, and EMA/ultrasound measures were then projected to a shared origin by centering, but without scaling in order to retain dimension-specific variation in movement range. The ultrasound data is noisy compared with the EMA data, so the ultrasound and EMA signals were both smoothed using the 5th-order Discrete Cosine Transform. See Figure \ref{fig:dlc_smooth} (right) for an example, which clearly shows the necessity of smoothing the ultrasound position data. Note that the EMA and ultrasound signals appear to capture the same underlying signal, but with a small time lag. This is likely a consequence of the EMA and ultrasound spatial points capturing slightly different locations on the tongue dorsum. Unfortunately, it is not straightforward to quantify specific relations between the EMA sensor locations and DLC knot locations, because the EMA sensors are not visible in the ultrasound image, so our selection of equivalent locations is a rough approximation.

\subsection{Feature extraction}

We analyse horizontal and vertical movements of the tongue dorsum (TD) and jaw (JAW). TD is a standard measurement dimension in the EMA literature, while JAW is an additional dimension that can be tracked using EMA and ultrasound. In the EMA data, TD is defined by the horizontal and vertical coordinates of the tongue dorsum sensor, while JAW is defined by the horizontal and vertical coordinates of the lower teeth sensor. In the ultrasound data, TD is the horizontal and vertical coordinates of DLC knot 5 and JAW is the mandibular short tendon knot \cite{strycharczuk-etal2025}. We use these knots as possible ultrasound correlates of TD and JAW, but it was not possible to verify that these represent identical physical locations as the EMA sensors. Our analysis instead focuses on how each signal captures the relative distances between vowels, rather than raw comparisons. Position and velocity trajectories were segmented into separate gestures, defined as an interval bounded by two zero-crossings in the velocity signal. Diphthongs and some long monophthongs can have two distinct velocity peaks \cite{strycharczuk-etal2024} and we also included closure and release gestures. We only retained trajectories for which there exists a matching EMA/ultrasound pair within a given sensor/dimension (e.g. TDx). In total, we analyse 2093 trajectories (630 TDx, 549 TDy, 504 JAWx, 410 JAWy). The different trajectory counts for x/y dimensions are a consequence of their different movement dynamics, which is not an issue for our present analysis, where we compare parameter estimation separately within dimensions.

\subsection{Parameter estimation and evaluation}

We estimate the coefficients for the parameters $b, k, T$ of a linear harmonic oscillator

\begin{equation}
\ddot{x} + b\dot{x} + k(x-T) = 0
\label{eq:lho}
\end{equation}

using constrained least squares. We optimize over the generic objective function in (\ref{sr3}), where $\dot{X}$ is a time series of derivatives, $\Theta(X)$ is a feature library comprised of the model parameters in (\ref{eq:lho}), and $\Xi$ is the coefficient matrix to be optimized.

\begin{equation}
\min\limits_{\Xi} \frac{1}{2}||\dot{X} - \Theta(X)\Xi||^{2} \text{\ \ \ subject to } C\xi = d
\label{sr3}
\end{equation}

We specifically solve for the acceleration of the system and integrate to obtain position and velocity trajectories that are evaluated against empirical data. We split the second-order differential equation into two first-order equations (1) $y = \dot{x}$; (2) $\dot{y} = -by -kx$, with the first equation subject to the linear constraint $y \mbeq 1.0\dot{x}$ to reduce model complexity \cite{champion-etal2020}. We use a maximum of 30 iterations to allow convergence of the optimization algorithm. After discovering optimal coefficients, we generate a simulated trajectory by solving a linear harmonic oscillator using the discovered coefficients and quantify fit between the modelled and empirical trajectories using $R^2$ values. This essentially follows the same process as in \citet{kirkham2025b}, but without any thresholding parameters, meaning that all model terms are used in fitting to data.

\section{Results}

\subsection{Model fit and parameter comparisons}

\begin{table}[th]
  \centering
  \begin{minipage}[t]{0.48\textwidth}
    \caption{$R^2$ model fit statistics for each variable, which summarizes the accuracy of model fits to empirical data.}
    \label{tab:fit}
    \centering
    \begin{tabular}{llrrrr}
      \toprule
      Variable & Modality & N & $\bar{R}^2$ & $R^2 \sigma$ & $R^2$ min, max\\
      \midrule
      TDx & EMA & 630 & 0.95 & 0.09 & 0.23, 1.00\\
       & US & 630 & 0.95 & 0.09 & 0.32, 1.00\\
      \midrule
      TDy & EMA & 549 & 0.94 & 0.09 & 0.38, 1.00\\
       &US & 549 & 0.95 & 0.10 & 0.37, 1.00\\
      \midrule
      JAWx & EMA & 504 & 0.90 & 0.15 & 0.27, 1.00\\
       & US & 504 & 0.93 & 0.13 & 0.29, 1.00\\
      \midrule
      JAWy & EMA & 410 & 0.92 & 0.12 & 0.36, 1.00\\
       & US & 410 & 0.93 & 0.12 & 0.32, 1.00\\
      \bottomrule
    \end{tabular}
  \end{minipage}
  \hfill
  \begin{minipage}[t]{0.48\textwidth}
    \caption{Bayesian mean and 95\% CIs for each parameter, which represents how much the ultrasound-estimated parameters deviate from the EMA-estimated parameters.}
    \label{tab:bayes}
    \centering
    \begin{tabular}{llrr}
      \toprule
      Parameter & Variable & $\bar{\beta}$ & 95\% CI\\
      \midrule
      $T$	& TDx & $-1.20$ & [$-2.35$, $-0.06$]\\
      			& TDy & $0.29$ & [$-0.32$, $+0.92$]\\
  	    		& JAWx & $-0.45$ & [$-0.74$, $-0.17$]\\
      			& JAWy & $-1.36$ & [$-1.81$, $-0.91$]\\
      \midrule
      $k$	& TDx & $-0.49$ & [$-3.96$, $+2.74$]\\
  		    	& TDy & $0.86$ & [$-2.38$, $+4.13$]\\
      				& JAWx & $2.29$ & [$-0.81$, $+5.22$]\\
  		    	& JAWy & $0.38$ & [$-2.88$, $+3.69$]\\
      \midrule
      $b$	& TDx & $-0.11$ & [$-3.25$, $+3.07$]\\
  	    		& TDy & $0.07$ & [$-3.18$, $+3.29$]\\
      			& JAWx & $1.91$ & [$-1.21$, $+5.08$]\\
      			& JAWy & $0.13$ & [$-3.00$, $+3.34$]\\
      \bottomrule
    \end{tabular}
  \end{minipage}
\end{table}

\begin{figure}[t]
  \centering
	\includegraphics[scale=0.6]{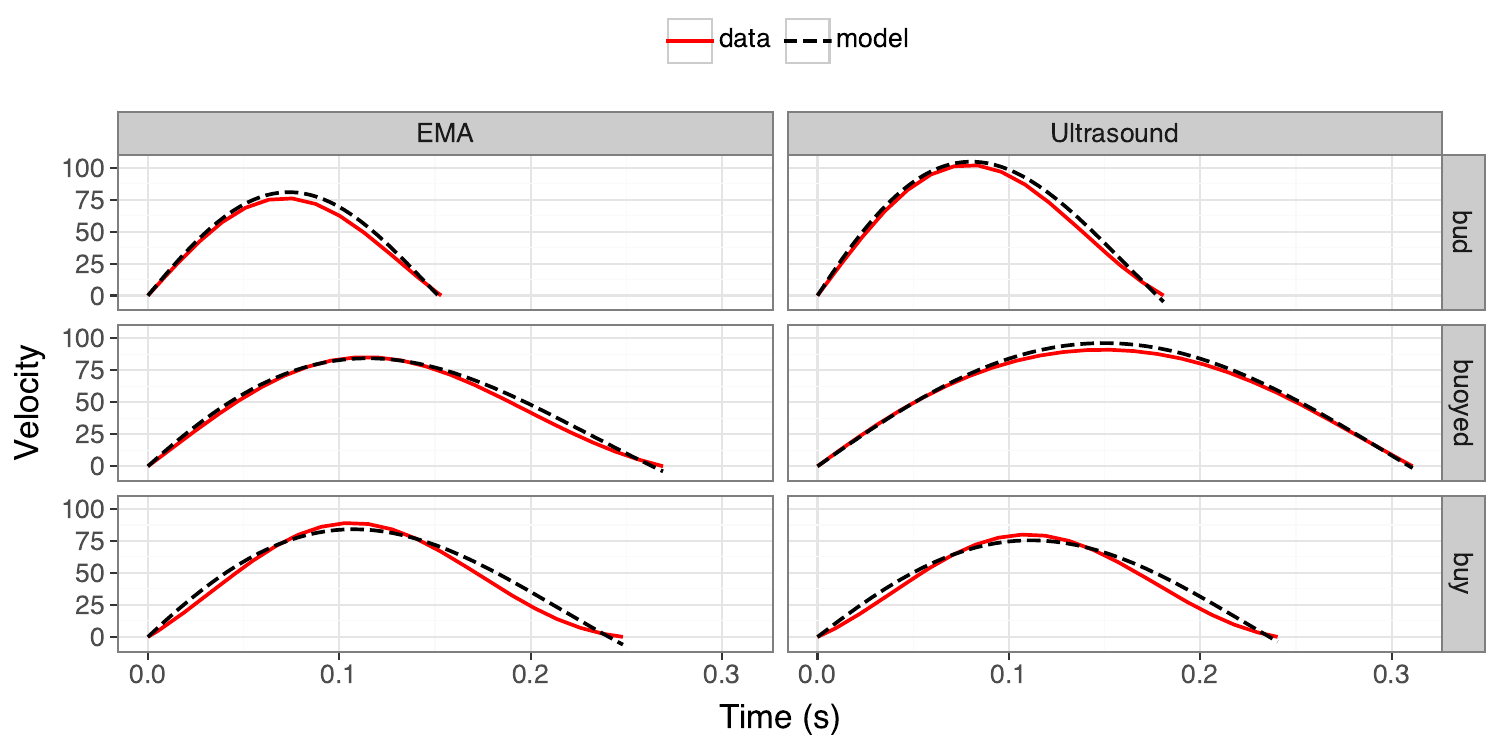}
  \caption{A random sample of three example velocity fits between EMA and Ultrasound for TDx. Tokens were selected using a fixed random seed and each word represents the same underlying token produced by a speaker. All fits are $R^2 > 0.92$.}
  \label{fig:model_fit_example}
\end{figure}

Table \ref{tab:fit} shows $R^2$ summary statistics for the fit between data and model predictions, with all variables at $R^2 \geq 0.9$. This suggests that good model fits can be achieved and that fitting accuracy is comparable between EMA and ultrasound, but that TD models fit slightly more accurately than JAW models. We visualize example velocity fits for TDx from each modality in Figure \ref{fig:model_fit_example}, which represents three tokens selected using a fixed random seed. It is apparent that the fits are qualitatively similar between EMA and ultrasound, with some small errors in the model predictions for each trajectory. We also note slight variation in the underlying data between modalities. This is likely to arise from similar sources as in Figure \ref{fig:dlc_smooth}, where we observe small time lags or slight durational differences.

Parameter values from EMA and ultrasound were then compared using the Bayesian hierarchical regression model: $y_{i} \sim \mathcal{N}(\alpha + \alpha_s[s_i] + (\beta + \beta_w[w_i]) \cdot \text{modality}_i, \sigma)$, where $y_{i}$ is an observation of the outcome variable, $\alpha$ is the intercept , $\beta$ is the effect of modality (EMA/ultrasound), $\beta_w \sim \mathcal{N}(0, \tau_\beta)$ is a by-word random slope for the effect of modality, $\alpha_s \sim \mathcal{N}(0, \tau_\alpha)$ is a speaker-level random intercept , and all other priors are weakly informative $\mathcal{N} \sim (0, 2)$. We ran MCMC sampling for 1000 warm-up iterations and 2000 sampling iterations using 4 chains, with the step size initialized at 0.1. Models were fitted using Stan v2.36 \cite{stan2024}. In all cases, EMA is the baseline variable, so the values represent how ultrasound differs from EMA.

Table \ref{tab:bayes} shows the mean effect of measurement modality on parameter estimation, along with 95\% credible intervals. In summary, when $\beta < 0$ it means that the ultrasound-estimated parameter is on average lower than the EMA-estimated parameter, whereas when  $\beta > 0$ the ultrasound-estimated parameter is on average higher than the EMA-estimated parameter. We find that the credible intervals cross zero across every variable for $k$ and $b$, suggesting no systematic difference between EMA and ultrasound in these parameters, largely due to a high degree of variability. The estimated $T$ difference is much narrower, where TDx, JAWx and JAWy have systematically lower estimated $T$ values in ultrasound (i.e. all 95\% CIs are below zero). The TDy ultrasound $T$ values are on average higher than the EMA values ($\bar{\beta} = 0.29$), but this is the one case for $T$ where the credible interval includes positive and negative values, indicating high uncertainty and no systematic differences.

\subsection{Word-specific differences}

We now investigate word-level effects to compare how the estimated parameters pattern between different words/vowels. This is important because EMA and ultrasound track points on the tongue in different ways, so we expect systematic effects of vowel height and anteriority. We visualize the difference between EMA and ultrasound modalities using the model's intercept and random slope coefficients, where EMA is the baseline. Note that each variable has a different range, so we focus on within-variable differences rather than between-variable comparisons.

\begin{figure}[t]
  \centering
  \includegraphics[scale=0.6]{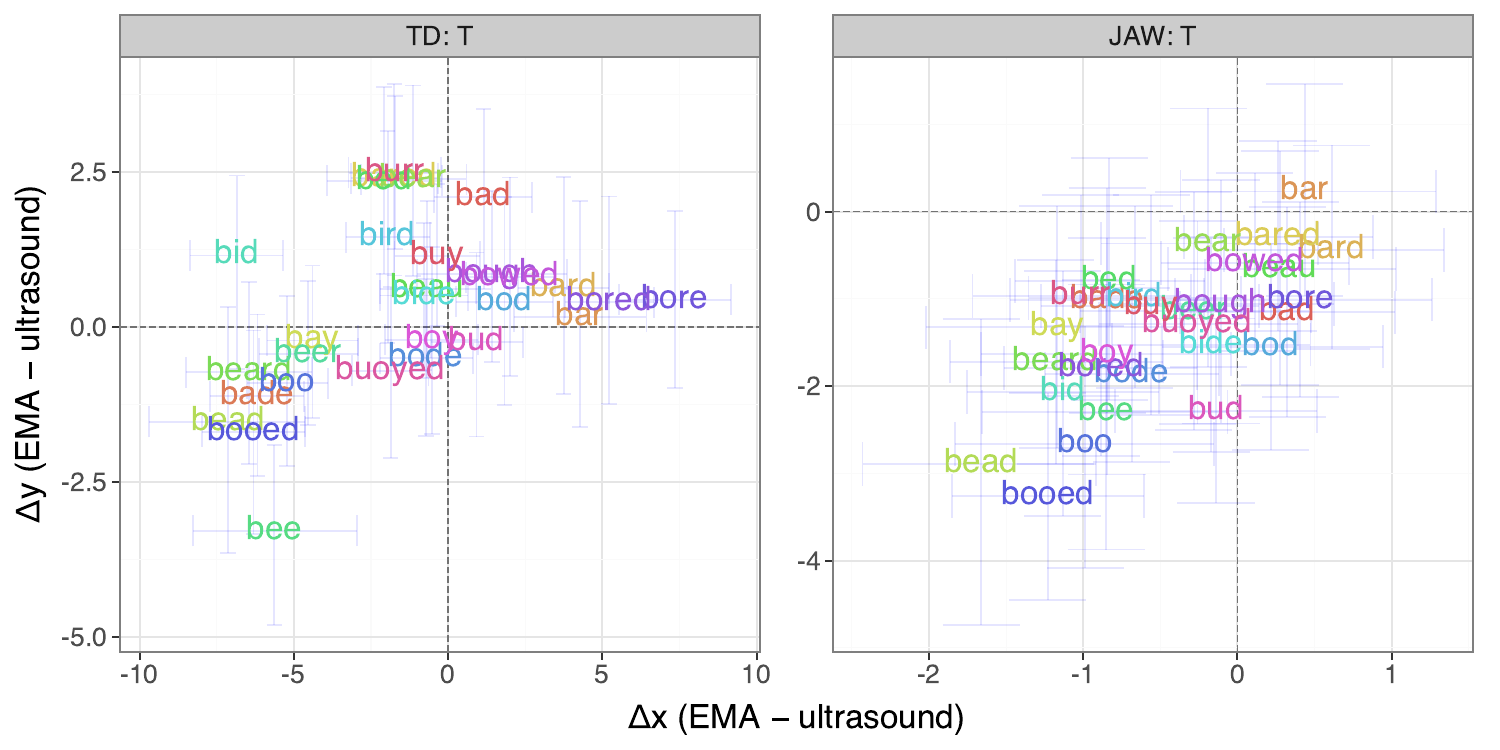}
  \caption{By-word effects showing how ultrasound-estimated $T$ (target) values differ from EMA-estimated values. Word labels show estimated means; blue lines show 95\% credible intervals. A value of zero indicates that ultrasound parameters do not differ from EMA parameters.}
  \label{fig:word_effects_T}
\end{figure}

Figure \ref{fig:word_effects_T} shows word-level effects for the target parameter $T$ in x/y space for TD and JAW, which represents the magnitude and direction of the difference between EMA and ultrasound. TD shows a systematic effect where front and high vowels, such as \textit{bee, bead, booed, beer}, have a lower and more posterior target for ultrasound parameters than EMA. Notably, these differences are consistent with previous research on how different ultrasound knots estimate vowel articulation, whereby the tongue dorsum knot underestimates the height and anteriority of front vowels \cite{strycharczuk-etal2025}. The results for JAW also show a systematic difference, although over a smaller range. Ultrasound underestimates the JAW target relative to EMA for front vowels (e.g. \textit{beed, booed, bid}) and overestimates it in some low vowels (e.g. \textit{bar)}. This is likely an artefact of probe stabilisation. In an ultrasound experiment, the probe is placed under the chin in its neutral position. When the jaw is lowered for the production of a low vowel, the soft tissue is squeezed against the probe, which can underestimate the distance between the short tendon and the probe, which leads to overestimation of the vertical JAW target.

\begin{figure}[t]
  \centering
  \includegraphics[scale=0.6]{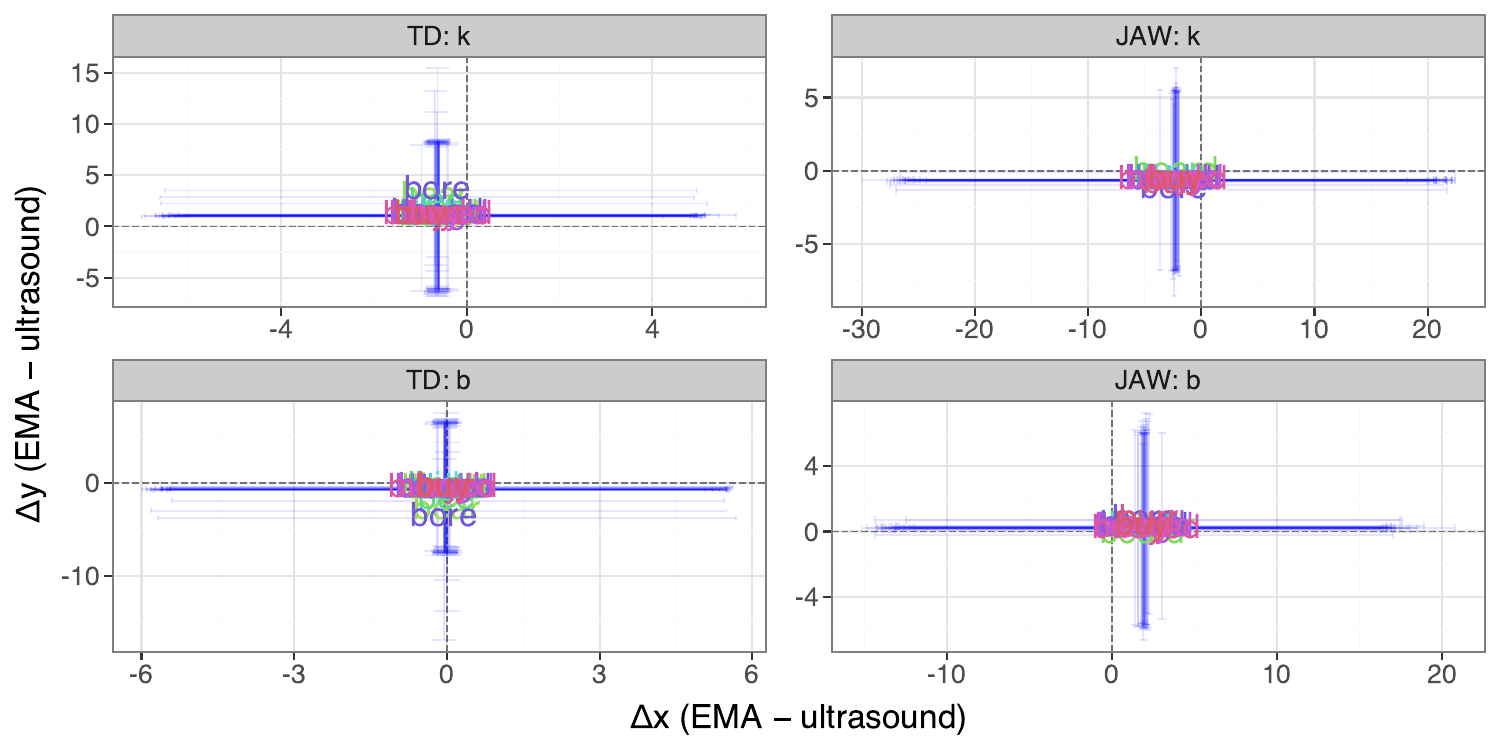}
  \caption{By-word effects showing how ultrasound-estimated $k$ (stiffness) and $b$ (damping) values differ from EMA-estimated values. Word labels show estimated means; blue lines show 95\% credible intervals. A value of zero indicates that ultrasound parameters do not differ from EMA parameters. Note that the word \emph{bar} has been removed from the JAW plots due to excessively large confidence intervals (crossing zero in both dimensions) that distort the axis ranges.}
  \label{fig:word_effects_bk}
\end{figure}

Figure \ref{fig:word_effects_bk} shows word-level random slope coefficients for the stiffness parameter $k$ and damping parameter $b$. In both cases, the majority of words cluster around zero with wide credible intervals. This indicates higher variability in measurement and a lack of systematic differences between EMA and ultrasound, except for a higher average $k$ and lower average $b$ in TDy for \textit{bore}. The JAW results show similar patterns, with near complete overlap, although note that \emph{bar} was removed from the JAW plots (but not the modelling) due to extremely wide credible intervals that skewed the plotting range. Note that variability in $k$ and $b$ estimation occurs in both EMA and ultrasound data, so it is not necessarily the case that only one modality produces extreme estimates. Overall, this suggests that estimation of $k$ and $b$ is not systematically different between EMA and ultrasound, but that estimates are much more variable than for the target ($T$) parameter.

\section{Discussion}

We estimated the parameters of a linear dynamical model from EMA- and ultrasound-derived kinematic measurements. The model is a reasonably good fit to EMA and ultrasound kinematics for the tongue dorsum and jaw, which shows that dynamical models can be fitted to ultrasound kinematic data with comparable accuracy to EMA data, at least in the case of British English vowels. Our second aim was to establish the nature of the estimated parameters. We do find differences between EMA and ultrasound parameters, but these differences are predictable based on known characteristics of how the ultrasound knot tracking captures tongue movement. For example, previous research shows that ultrasound knot 5 underestimates dorsum height in front vowels, but other knots (e.g. anterior knot 7) inaccurately estimate dorsum height in low vowels \cite{strycharczuk-etal2025}. A post-hoc comparison shows that the $T$ parameter estimated from TDy is more highly correlated with $T$ estimated from knot 7Y ($r = 0.83$) than from knot 5Y ($r = 0.60$), where knot 7 captures a more anterior part on the tongue. This is because knot 7Y more substantially \textit{overestimates} the target in low vowels in addition to the knot 5 high vowel differences, making the magnitude of EMA/ultrasound difference more linearly related to vowel height. In summary, while we find some differences in estimated tongue dorsum target parameters, these can be explained by the consequences of selecting a specific articulatory dimension to represent the tongue dorsum. Note also that EMA does not necessarily represent a golden standard in this respect, given flexibility in EMA sensor placement \cite{rebernik-etal2021}. Moreover, ultrasound offers additional opportunities in tracking multiple points from tongue tip to root. As such, it is of paramount importance to report the precise dimensions used for model fitting.

Flexibility of point selection is less relevant for JAW, where ultrasound only tracks a single point. Despite this, JAW targets also vary between EMA and ultrasound, but in a way that is consistent. Ultrasound underestimates JAW targets in high vowels and overestimates in low vowels compared with EMA. Importantly, this variation is not random and is clearly related to both probe movement and vowel height, suggesting that the mandibular short tendon tracked in the ultrasound image captures systematic aspects of jaw position, but in a different manner to the EMA lower teeth sensor. This goes some way towards validating short tendon tracking as capturing meaningful aspects of vowel articulation, which significantly expands the utility of ultrasound imaging beyond the tongue surface.

The ultrasound kinematic data are noisier than EMA and the underlying signal may be slightly obscured even when smoothed, which could be another source of differences. \citet{wrench-balch-tomes2022} analyse the same data we use in this study and show that horizontal tongue tip and blade measures correlate well between EMA sensors and DLC-tracked knots from ultrasound images ($r \geq 0.88$), but other articulatory dimensions have correlation coefficients $r < 0.4$. This suggests that EMA and ultrasound capture similar information in the tongue tip horizontal direction, but may capture differential spatial information for other dimensions. To this end, there is also scope for improving ultrasound kinematic measures, with an obvious area being better frame-to-frame tracking of DLC knots. At present, the DLC model estimates kinematics on a frame-by-frame basis, but inherent noise and measurement inaccuracy mean that knot tracking is unlikely to return a smooth function of time. One solution is to constrain tracking such that sharp divergences between frames are penalized. Finally, we focused on fitting a simple dynamical model to the data: an undamped linear harmonic oscillator. This allowed us to identify unique models for each token, with clearly interpretable parameters, but in future work it would be productive to compare more complex models, such as nonlinear dynamical models \cite{sorensen-gafos2016, stern-shaw2025}, as well as nonautonomous models with time-dependent parameters \cite{elie-etal2023}.

\section{Conclusions}

We show that a linear task dynamic model can be fitted to ultrasound kinematic data with a relatively high degree of accuracy, conditional on sufficient smoothing and segmentation. The estimated model parameters differ in specific ways between EMA and ultrasound, but the differences are systematic and are a consequence of the selected measurement dimensions. As a result, these results broaden the possibilities for lingual kinematic analysis, with ultrasound providing information on the tongue posterior, as well as a greater number of points along the tongue. We also find that tracking the mandibular short tendon allows for meaningful jaw movement dynamics, opening up new directions for the study of inter-articulator coordination in dynamical theories of speech production.

\section*{Author declarations}

The authors report funding from AHRC grants AH/S011900/1 (P.S., S.K.) and AH/Y002822/1 (S.K.), and British Academy grant MFSS24\textbackslash40076 (P.S.).

\section*{Ethics approval}

This study received ethical approval from Lancaster University Ethics Committee (Ref: FL18188) and the University of Manchester’s Proportionate University Research Ethics Committee (Ref: 2022-13946-22714).


\begin{thebibliography}{26}
\def\enquote#1{``#1,''}
\def\plainquote#1{``#1''}
\expandafter\ifx\csname natexlab\endcsname\relax\def\natexlab#1{#1}\fi
\providecommand{\dourl}[1]{\href{http://#1}{\nolinkurl{#1}}}
\providecommand{\bibinfo}[2]{#2}
\providecommand{\noopsort}[1]{}
\providecommand{\switchargs}[2]{#2#1}
  \def\eatspace #1{#1}

\bibitem[{Browman and Goldstein(1986)}]{browman-goldstein1986}
\bibinfo{author}{Browman, C.~P.},  and \bibinfo{author}{Goldstein, L.~M.}
  (\textbf{\bibinfo{year}{1986}}). \enquote{\bibinfo{title}{Towards an
  articulatory phonology}} \bibinfo{journal}{Phonology} \textbf{3}(1),
  \bibinfo{pages}{219--252}.

\bibitem[{Byrd and Saltzman(1998)}]{byrd-saltzman1998}
\bibinfo{author}{Byrd, D.},  and \bibinfo{author}{Saltzman, E.}
  (\textbf{\bibinfo{year}{1998}}). \enquote{\bibinfo{title}{Intragestural
  dynamics of multiple prosodic boundaries}} \bibinfo{journal}{Journal of
  Phonetics} \textbf{26}(2), \bibinfo{pages}{173--199}.

\bibitem[{Champion \emph{et~al.}(2020)Champion, Zheng, Aravkin, Brunton, and
  Kutz}]{champion-etal2020}
\bibinfo{author}{Champion, K.}, \bibinfo{author}{Zheng, P.},
  \bibinfo{author}{Aravkin, A.~Y.}, \bibinfo{author}{Brunton, S.~L.},  and
  \bibinfo{author}{Kutz, J.~N.} (\textbf{\bibinfo{year}{2020}}).
  \enquote{\bibinfo{title}{A unified sparse optimization framework to learn
  parsimonious physics-informed models from data}} \bibinfo{journal}{{IEEE}
  Access} \textbf{8}, \bibinfo{pages}{169259--169271}.

\bibitem[{Elie \emph{et~al.}(2023)Elie, Lee, and Turk}]{elie-etal2023}
\bibinfo{author}{Elie, B.}, \bibinfo{author}{Lee, D.~N.},  and
  \bibinfo{author}{Turk, A.} (\textbf{\bibinfo{year}{2023}}).
  \enquote{\bibinfo{title}{Modeling trajectories of human speech articulators
  using general {T}au theory}} \bibinfo{journal}{Speech Communication}
  \textbf{151}, \bibinfo{pages}{24--38}.

\bibitem[{Fowler(1980)}]{fowler1980}
\bibinfo{author}{Fowler, C.~A.} (\textbf{\bibinfo{year}{1980}}).
  \enquote{\bibinfo{title}{Coarticulation and theories of extrinsic timing}}
  \bibinfo{journal}{Journal of Phonetics} \textbf{8}(1),
  \bibinfo{pages}{113--133}.

\bibitem[{Gafos(2006)}]{gafos2006}
\bibinfo{author}{Gafos, A.~I.} (\textbf{\bibinfo{year}{2006}}).
  \enquote{\bibinfo{title}{Dynamics in grammar}} in
  \emph{\bibinfo{booktitle}{Laboratory Phonology 8: Varieties of Phonological
  Competence}}, edited by \bibinfo{editor}{L.~Goldstein},
  \bibinfo{editor}{D.~Whalen}, and \bibinfo{editor}{C.~T. Best}
  (\bibinfo{publisher}{Mouton de Gruyter}, \bibinfo{address}{Berlin}), pp.
  \bibinfo{pages}{51--79}.

\bibitem[{Iskarous(2016)}]{iskarous2016}
\bibinfo{author}{Iskarous, K.} (\textbf{\bibinfo{year}{2016}}).
  \enquote{\bibinfo{title}{Compatible dynamical models of environmental,
  sensory, and perceptual systems}} \bibinfo{journal}{Ecological Psychology}
  \textbf{28}(4), \bibinfo{pages}{295--311}.

\bibitem[{Kirkham(2024)}]{kirkham2024}
\bibinfo{author}{Kirkham, S.} (\textbf{\bibinfo{year}{2024}}).
  \enquote{\bibinfo{title}{Discovering dynamical models of speech using
  physics-informed machine learning}} \bibinfo{journal}{Proc. ISSP 2024 -- 13th
  International Seminar on Speech Production} \bibinfo{pages}{185--188}.

\bibitem[{Kirkham(2025{\natexlab{a}})}]{kirkham2025b}
\bibinfo{author}{Kirkham, S.} (\textbf{\bibinfo{year}{2025}}{\natexlab{a}}).
  \enquote{\bibinfo{title}{Discovering dynamical laws for speech gestures}}
  \bibinfo{journal}{Cognitive Science} \textbf{49}(5), \bibinfo{pages}{e70064}.

\bibitem[{Kirkham(2025{\natexlab{b}})}]{kirkham2025}
\bibinfo{author}{Kirkham, S.} (\textbf{\bibinfo{year}{2025}}{\natexlab{b}}).
  \enquote{\bibinfo{title}{Scaling laws for nonlinear dynamical models of
  articulatory control}} \bibinfo{journal}{{JASA} Express Letters}
  \textbf{5}(2), \bibinfo{pages}{1--7}.

\bibitem[{Kirkham \emph{et~al.}(2023)Kirkham, Strycharczuk, Gorman, Nagamine,
  and Wrench}]{kirkham-etal2023}
\bibinfo{author}{Kirkham, S.}, \bibinfo{author}{Strycharczuk, P.},
  \bibinfo{author}{Gorman, E.}, \bibinfo{author}{Nagamine, T.},  and
  \bibinfo{author}{Wrench, A.} (\textbf{\bibinfo{year}{2023}}).
  \enquote{\bibinfo{title}{Co-registration of simultaneous high-speed
  ultrasound and electromagnetic articulography for speech production
  research}} \bibinfo{journal}{Proceedings of the 19th International Congress
  of Phonetic Sciences} \bibinfo{pages}{942--946}.

\bibitem[{Lammert \emph{et~al.}(2013)Lammert, Goldstein, Narayanan, and
  Iskarous}]{lammert-etal2013b}
\bibinfo{author}{Lammert, A.}, \bibinfo{author}{Goldstein, L.},
  \bibinfo{author}{Narayanan, S.},  and \bibinfo{author}{Iskarous, K.}
  (\textbf{\bibinfo{year}{2013}}). \enquote{\bibinfo{title}{Statistical methods
  for estimation of direct and differential kinematics of the vocal tract}}
  \bibinfo{journal}{Speech Communication} \textbf{55}(1),
  \bibinfo{pages}{147--161}.

\bibitem[{Mathis \emph{et~al.}(2018)Mathis, Mamidanna, Mury, Abe, Murthy,
  Mathis, and Bethge}]{mathis-etal2018}
\bibinfo{author}{Mathis, A.}, \bibinfo{author}{Mamidanna, P.},
  \bibinfo{author}{Mury, K.~M.}, \bibinfo{author}{Abe, T.},
  \bibinfo{author}{Murthy, V.~N.}, \bibinfo{author}{Mathis, M.~W.},  and
  \bibinfo{author}{Bethge, M.} (\textbf{\bibinfo{year}{2018}}).
  \enquote{\bibinfo{title}{{DeepLabCut}: Markerless pose estimation of
  user-defined body parts with deep learning}} \bibinfo{journal}{Nature
  Neuroscience} \textbf{21}(9), \bibinfo{pages}{1281--1289}.

\bibitem[{McAuliffe \emph{et~al.}(2017)McAuliffe, Socolof, Mihuc, Wagner, and
  Sonderegger}]{mcauliffe-etal2017}
\bibinfo{author}{McAuliffe, M.}, \bibinfo{author}{Socolof, M.},
  \bibinfo{author}{Mihuc, S.}, \bibinfo{author}{Wagner, M.},  and
  \bibinfo{author}{Sonderegger, M.} (\textbf{\bibinfo{year}{2017}}).
  \enquote{\bibinfo{title}{{Montreal Forced Aligner}: Trainable text-speech
  alignment using {Kaldi}}} in \emph{\bibinfo{booktitle}{Proc. {Interspeech}
  2017}}, pp. \bibinfo{pages}{498--502}.

\bibitem[{Rebernik \emph{et~al.}(2021)Rebernik, Jacobi, Jonkers, Noiray, and
  Wieling}]{rebernik-etal2021}
\bibinfo{author}{Rebernik, T.}, \bibinfo{author}{Jacobi, J.},
  \bibinfo{author}{Jonkers, R.}, \bibinfo{author}{Noiray, A.},  and
  \bibinfo{author}{Wieling, M.} (\textbf{\bibinfo{year}{2021}}).
  \enquote{\bibinfo{title}{A review of data collection practices using
  electromagnetic articulograph}} \bibinfo{journal}{Laboratory Phonology}
  \textbf{12}(1), \bibinfo{pages}{1--42}.

\bibitem[{Saltzman and Munhall(1989)}]{saltzman-munhall1989}
\bibinfo{author}{Saltzman, E.},  and \bibinfo{author}{Munhall, K.~G.}
  (\textbf{\bibinfo{year}{1989}}). \enquote{\bibinfo{title}{A dynamical
  approach to gestural patterning in speech production}}
  \bibinfo{journal}{Ecological Psychology} \textbf{1}(4),
  \bibinfo{pages}{333--382}.

\bibitem[{Sorensen and Gafos(2016)}]{sorensen-gafos2016}
\bibinfo{author}{Sorensen, T.},  and \bibinfo{author}{Gafos, A.~I.}
  (\textbf{\bibinfo{year}{2016}}). \enquote{\bibinfo{title}{The gesture as an
  autonomous nonlinear dynamical system}} \bibinfo{journal}{Ecological
  Psychology} \textbf{28}(4), \bibinfo{pages}{188--215}.

\bibitem[{Spreafico \emph{et~al.}(2018)Spreafico, Pucher, and
  Matosova}]{spreafico-eta2018}
\bibinfo{author}{Spreafico, L.}, \bibinfo{author}{Pucher, M.},  and
  \bibinfo{author}{Matosova, A.} (\textbf{\bibinfo{year}{2018}}).
  \enquote{\bibinfo{title}{{UltraFit}: A speaker-friendly headset for
  ultrasound recordings in speech science}} \bibinfo{journal}{Proceedings of
  Interspeech 2018} \bibinfo{pages}{1--4}.

\bibitem[{{Stan Development Team}(2024)}]{stan2024}
\bibinfo{author}{{Stan Development Team}} (\textbf{\bibinfo{year}{2024}}).
  \plainquote{\bibinfo{title}{{Stan Reference Manual}, v2.36.0}}
  \bibinfo{howpublished}{https://mc-stan.org}.

\bibitem[{Stern and Shaw(2025)}]{stern-shaw2025}
\bibinfo{author}{Stern, M.~C.},  and \bibinfo{author}{Shaw, J.~A.}
  (\textbf{\bibinfo{year}{2025}}). \enquote{\bibinfo{title}{Nonlinear
  second-order dynamics describe labial constriction trajectories across
  languages and contexts}} \bibinfo{journal}{Journal of Phonetics}
  \textbf{111}(101427), \bibinfo{pages}{1--18}.

\bibitem[{Strycharczuk \emph{et~al.}(2024)Strycharczuk, Kirkham, Gorman, and
  Nagamine}]{strycharczuk-etal2024}
\bibinfo{author}{Strycharczuk, P.}, \bibinfo{author}{Kirkham, S.},
  \bibinfo{author}{Gorman, E.},  and \bibinfo{author}{Nagamine, T.}
  (\textbf{\bibinfo{year}{2024}}). \enquote{\bibinfo{title}{Towards a dynamical
  model of {E}nglish vowels: Evidence from diphthongisation}}
  \bibinfo{journal}{Journal of Phonetics} \textbf{107}, \bibinfo{pages}{1--26}.

\bibitem[{Strycharczuk \emph{et~al.}(2025)Strycharczuk, Kirkham, Gorman, and
  Nagamine}]{strycharczuk-etal2025}
\bibinfo{author}{Strycharczuk, P.}, \bibinfo{author}{Kirkham, S.},
  \bibinfo{author}{Gorman, E.},  and \bibinfo{author}{Nagamine, T.}
  (\textbf{\bibinfo{year}{2025}}). \enquote{\bibinfo{title}{Dimensionality
  reduction in lingual articulation of vowels: Evidence from lax vowels in
  {N}orthern {A}nglo-{E}nglish}} \bibinfo{journal}{Language and Speech} .

\bibitem[{Strycharczuk and Scobbie(2015)}]{strycharczuk-scobbie2015}
\bibinfo{author}{Strycharczuk, P.},  and \bibinfo{author}{Scobbie, J.~M.}
  (\textbf{\bibinfo{year}{2015}}). \enquote{\bibinfo{title}{Velocity measures
  in ultrasound data. gestural timing of post-vocalic /l/ in {E}nglish}}
  \bibinfo{journal}{Proceedings of the XVIII International Congress of Phonetic
  Sciences} \bibinfo{pages}{1--5}.

\bibitem[{Tilsen(2016)}]{tilsen2016}
\bibinfo{author}{Tilsen, S.} (\textbf{\bibinfo{year}{2016}}).
  \enquote{\bibinfo{title}{Selection and coordination: The articulatory basis
  for the emergence of phonological structure}} \bibinfo{journal}{Journal of
  Phonetics} \textbf{55}, \bibinfo{pages}{53--77}.

\bibitem[{Wrench(2022)}]{aaa2022}
\bibinfo{author}{Wrench, A.} (\textbf{\bibinfo{year}{2022}}).
  \plainquote{\bibinfo{title}{{Articulate Assistant Advanced} v.220.04
  [software]}} .

\bibitem[{Wrench and Balch-Tomes(2022)}]{wrench-balch-tomes2022}
\bibinfo{author}{Wrench, A.},  and \bibinfo{author}{Balch-Tomes, J.}
  (\textbf{\bibinfo{year}{2022}}). \enquote{\bibinfo{title}{Beyond the edge:
  Markerless pose estimation of speech articulators from ultrasound and camera
  images using {DeepLabCut}}} \bibinfo{journal}{Sensors} \textbf{22}(1133),
  \bibinfo{pages}{1--29}.

\end{thebibliography}
\end{document}